\documentclass[10pt,journal]{IEEEtran}
\usepackage{amsmath,amsfonts}
\usepackage{algorithmic}
\usepackage{array}
\usepackage{soul}
\usepackage{threeparttable}
\usepackage{color}
\usepackage{tikz}
\usepackage{pgfplots}
\usepackage{float}
\usepackage{textcomp}
\usepackage{stfloats}
\usepackage{url}
\usepackage{verbatim}
\usepackage{graphicx}
\usepackage{subcaption}
\usepackage{cite}
\usepackage[linesnumbered,algoruled,vlined]{algorithm2e}
\usepackage{booktabs}
\usepackage{multirow}
\usepackage{makecell}
\usepackage{longtable}

\usepackage{amsthm}
\theoremstyle{plain}

\usepackage{mathtools}

\DeclarePairedDelimiterX{\infdivx}[2]{(}{)}{%
  #1\;\delimsize\|\;#2%
}

\pgfplotsset{compat=1.18} 
\usepackage{environ}
\NewEnviron{myequation}{
    \begin{equation*}
    \scalebox{1}{$\BODY$}
    \end{equation*}
    }

\begin{document}

\title{Double Oracle Neural Architecture Search for Game Theoretic Deep Learning Models}

%author{IEEE Publication Technology,~\IEEEmembership{Staff,~IEEE,}

\author{Aye Phyu Phyu Aung, Xinrun Wang, Ruiyu Wang, Hau Chan, Bo An, Xiaoli Li \IEEEmembership{Fellow,~IEEE}, ~J.~Senthilnath,~\IEEEmembership{Senior~Member,~IEEE}\\

%{\tt\small \{aye\_phyu\_phyu\_aung, xlli, j\_senthilnath\}@i2r.a-star.edu.sg}, \\ {\tt \small \{xinrun.wang, boan\}@ntu.edu.sg, ruiyuw@kth.se, hchan3@unl.edu}\\

%{\tt\small \{ayep0001,xinrun.wang, boan\}@ntu.edu.sg, runshengyu@gmail.com,}\\ {\tt\small \{j\_senthilnath, xlli\}@i2r.a-star.edu.sg}\\
        % <-this % stops a space
% \thanks{This paper was produced by the IEEE Publication Technology Group. They are in Piscataway, NJ.}% <-this % stops a space
\thanks{Manuscript received April 4, 2024}
\thanks{\noindent Aye Phyu Phyu Aung, and J.Senthilnath are with the Institute for Infocomm Research, Agency for Science, Technology and Research (A*STAR), Singapore. \{email: aye\_phyu\_phyu\_aung,  J\_Senthilnath@i2r.a-star.edu.sg\} }
\thanks{Xinrun Wang, and Bo An are With the School of Computer Science and Engineering, Nanyang Technological University, Singapore. \{email: xinrun.wang, boan@ntu.edu.sg\}} 
\thanks{Ruiyu Wang is with KTH Royal Institute of Technology, \{email: ruiyuw@kth.se\}}
\thanks {Hau Chan is with the University of Nebraska-Lincoln, \{email: hchan3@unl.edu\} }
\thanks{Xiaoli Li is with the Institute for Infocomm Research, Agency for Science, Technology and Research (A*STAR), Singapore, School of Computer Science and Engineering, Nanyang Technological University, Singapore and A*STAR Centre for Frontier AI Research, Singapore. \{email: xlli@i2r.a-star.edu.sg\}} 
\thanks{Aye Phyu Phyu Aung and Xinrun Wang equally contributed to the paper. Aye Phyu Phyu Aung, Xinrun Wang, and J. Senthilnath serve as corresponding authors.}
\thanks{This paper is the extension of our preliminary work "DO-GAN: A Double Oracle Framework for Generative Adversarial Networks" appeared in CVPR,2022. Also, this study partly appeared on Chapter 6 of Aye Phyu Phyu Aung's thesis dissertation.}
%However, it has not been published in any journal or conference.
}

% The paper headers
\markboth{}%
{Shell \MakeLowercase{\textit{et al.}}: A Sample Article Using IEEEtran.cls for IEEE Journals}

%\IEEEpubid{0000--0000/00\$00.00~\copyright~2024 IEEE}
% Remember, if you use this you must call \IEEEpubidadjcol in the second
% column for its text to clear the IEEEpubid mark.

\maketitle

\begin{abstract}
In this paper, we propose a new approach to train deep learning models using game theory concepts including Generative Adversarial Networks (GANs) and Adversarial Training (AT) where we deploy a double-oracle framework using best response oracles. GAN is essentially a two-player zero-sum game between the generator and the discriminator. The same concept can be applied to AT with attacker and classifier as players. Training these models is challenging as a pure Nash equilibrium may not exist and even finding the mixed Nash equilibrium is difficult as training algorithms for both GAN and AT have a large-scale strategy space. Extending our preliminary model DO-GAN, we propose the methods to apply the double oracle framework concept to Adversarial Neural Architecture Search (NAS for GAN) and Adversarial Training (NAS for AT) algorithms. We first generalize the players' strategies as the trained models of generator and discriminator from the best response oracles. We then compute the meta-strategies using a linear program. For scalability of the framework where multiple network models of best responses are stored in the memory, we prune the weakly-dominated players' strategies to keep the oracles from becoming intractable. Finally, we conduct experiments on MNIST, CIFAR-10 and TinyImageNet for DONAS-GAN. We also evaluate the robustness under FGSM and PGD attacks on CIFAR-10, SVHN and TinyImageNet for DONAS-AT. We show that all our variants have significant improvements in both subjective qualitative evaluation and quantitative metrics, compared with their respective base architectures.
\end{abstract}


\newpage
\section{Introduction}
\IEEEPARstart{M}{ost} machine learning algorithms involve optimizing a single set of parameters to minimize %decrease 
a single cost function. For some deep learning models like GAN and AT, however, two or more ``players'' adapt their own parameters to minimize %decrease 
their own cost, in competition with the other players like a game. There are two most popular architectures that apply game theory concepts in deep learning models: i) generative adversarial networks (GAN) where a generator network generates images, while the discriminator distinguishes the generated images from real images, resulting in the ability of the generator to produce realistic images~\cite{goodfellow2014generative,brock2018large,karras2019style}, and ii) adversarial training (AT) where a classifier is trained against an attacker who can manipulate the inputs to decrease the performance of the classifier~\cite{goodfellow2014explaining,madry2017towards}. In both cases, the training algorithms can be viewed as a two-player zero-sum game where each player is alternately trained to maximize their respective utilities until convergence corresponding
to a Nash Equilibrium (NE)~\cite{arora2017generalization}.

However, pure NE cannot be reliably reached by existing algorithms as pure NE may not exist ~\cite{farnia2020gans,mescheder2017numerics}. This also leads to unstable training in GANs depending on the data and the hyperparameters. In this case, mixed NE is a more suitable solution concept~\cite{hsieh2018finding}. Several recent works propose mixture architectures with multiple generators and discriminators that consider mixed NE such as MIX+GAN~\cite{arora2017generalization} and MGAN~\cite{hoang2018mgan} but theoretically cannot guarantee to converge to mixed NE. Mirror-GAN~\cite{hsieh2018finding} computes the mixed NE by sampling over the infinite-dimensional strategy space and proposes provably convergent proximal methods. However, the sampling approach may not be efficient as mixed NE may only have a few strategies in the support set. 

On the other hand, Neural Architecture Search (NAS) approaches have also shown promising results. %The NAS technique has 
They have been applied to some computer vision tasks, such as image classification, dense image prediction and object detection. 
Recently, NAS has been researched in GANs~\cite{gong2019autogan,rezaei2021generative}. Particularly, Adversarial-NAS~\cite{gao2020adversarialnas} searches both of the generator and discriminator simultaneously in a differentiable manner using gradient-based method NAS and a large architecture search space. Moreover, NAS is also used to search the classifier architectures in Adversarial Training (AT) for robustness. AdvRush~\cite{mok2021advrush} considers both white-box attacks and black-box attacks using the input loss landscape of the networks to represent their intrinsic robustness. However, the NAS techniques only derive one final architecture with a criterion such as maximum weight from the search space, ignoring other top-performing results from the neural architecture search.

\IEEEpubidadjcol

Double Oracle (DO) algorithm~\cite{mcmahan2003planning} is a powerful framework to compute mixed NE in large-scale games.  
The algorithm starts with a restricted game that is initialized with a small set of actions and solves it to get the NE strategies of the restricted game. The algorithm then computes players' best responses using oracles to the NE strategies and adds them into the restricted game for the next iteration.
DO framework has been applied in various disciplines~\cite{jain2011double,bosansky2013double}, as well as Multi-agent Reinforcement Learning (MARL)~\cite{lanctot2017unified}. Our previous work DO-GAN~\cite{aung2021gan} train GANs by deploying a double oracle framework using generator and discriminator from the best response oracles. 

Extending the successful applications of the DO framework on GAN, we propose a Double Oracle Neural Architecture Search for GAN and AT (DONAS-GAN and DONAS-AT) to show the improvement of performance by the double oracle framework on neural architecture search for the two game theoretic deep learning models. This paper presents four key contributions. 

\begin{enumerate}
    \item We propose the general double oracle framework for Neural Architecture Search (DONAS) with two main components of GAN and AT (generator/discriminator or classifier/attacker) as players. The players in DONAS obtain the best responses from their oracles and add the utilities to a meta-matrix. 
    We use a linear program to obtain the probability distributions of the players' pure strategies (meta-strategies) for the respective oracles and pruning for scalable implementation of the extended algorithms. 
    \item We propose DONAS-GAN variant with the respective oracles to search multiple generators and discriminators referencing the techniques from Adversarial-NAS. 
    \item We also propose DONAS-AT variant with the classifier/attacker oracles for robustness in the classification tasks referencing the techniques from AdvRush to search the classifier model and applying the general double oracle framework.
    \item We propose two methods: (a): Harmonic Mean and (b): Nash, to sequentially finetune the searched networks generators and discriminators for DONAS-GAN as well as classifiers for DONAS-AT.
\end{enumerate}
 Finally, we provide a comprehensive evaluation on the performances of DONAS-GAN and DONAS-AT using real-world datasets. Experiment results show that our variants have significant improvements in terms of both subjective qualitative evaluation and quantitative metrics i.e., FID score.

\section{Related Works}
In this section, we briefly introduce existing game theoretic deep learning architectures, double oracle algorithm and its applications that are related to our work. 

\subsection{GAN, AT and NAS}
\noindent \textbf{Generative Adversarial Networks (GAN) Architectures.}
Various GAN architectures have been proposed to improve their performance. 
Apart from single architecture advancements, multi-model architectures have also shown promising improvements to the GAN training process.
Considering mixed NE, MIX+GAN~\cite{arora2017generalization} maintains a mixture of generators and discriminators with the same network architecture but has its own trainable parameters. However, training a mixture of networks without parameter sharing makes the algorithm computationally expensive. Mixture Generative Adversarial Nets (MGAN)~\cite{hoang2018mgan} proposes to capture diverse data modes by formulating GAN as a game between a classifier, a discriminator and multiple generators with parameter sharing. However, MIX+GAN and MGAN cannot converge to mixed NE. Mirror-GAN~\cite{hsieh2018finding} finds the mixed NE by sampling over the infinite-dimensional strategy space and proposes provably convergent proximal methods. This approach may be inefficient to compute mixed NE as the mixed NE may only have a few strategies with positive probabilities in the infinite strategy space.  \cite{gao2020adversarialnas}

\noindent \textbf{Adversarial Training (AT).} As the NAS becomes widely researched, investigations on its intrinsic vulnerability and enhancement of robustness have also increased. Numerous approaches have been proposed but Adversarial Training (AT) based methods~\cite{madry2017towards} still remain the strongest defense approach. RobNet~\cite{guo2020meets}, randomly samples architectures from a search space and adversarially trains each of them. However, this is very computationally expensive. Moreover, evolutionary algorithm architectures such as %R-NAS~\cite{kotyan2020towards} and 
RoCo-NAS~\cite{geraeinejad2021roco} are restricted to only black-box attacks. 

\noindent \textbf{Neural Architecture Search (NAS).} With Automatic Machine Learning (AutoML), Neural Architecture Search (NAS) has become one of the most important directions for machine learning. 
%The goal of NAS is to automatically search for an effective architecture that satisfies certain demands. The DARTS in one-shot literature is the first approach that relaxes the search space to be continuous and conducts searching in a differentiable way. The architecture parameters and network weights can be trained simultaneously\cite{liu2018darts}. However, they are extremely time-consuming to obtain the generators. 
In the context of generative networks, AutoGAN~\cite{gong2019autogan} and Adversarial-NAS~\cite{gao2020adversarialnas} search the generator and discriminator architectures simultaneously in a differentiable manner and  GA-NAS~\cite{rezaei2021generative} proposes to use adversarial learning approach where the searched generator is trained by reinforcement learning based on rewards provided by 
 the discriminator. Similarly for AT, AdvRush~\cite{mok2021advrush} proposes an adversarial robustness-aware neural architecture search algorithm using the input loss landscape of the neural networks to represent their intrinsic robustness. However, in both GAN and AT, NAS only samples the architectures from the search space by picking the operations with maximum weights. To improve the modes that are missed for training, we propose DONAS for GANs and AT by sampling and finetuning multiple networks instead of just selecting the maximum. DONAS does not induce any restriction on NAS.

\subsection{Double Oracle Algorithm.}
Double Oracle (DO) algorithm starts with a small restricted game between two players and solves it to get the players' strategies at Nash Equilibrium (NE) of the restricted game. The algorithm then exploits the respective best response oracles for additional strategies of the players. The DO algorithm terminates when the best response utilities are not higher than the equilibrium utility of the current restricted game. Hence, it finds the NE of the game without enumerating the entire strategy space. Moreover, in two-player zero-sum games, DO converges to a min-max equilibrium~\cite{mcmahan2003planning}. DO framework is used to solve large-scale normal-form and extensive-form games such as security games~\cite{tsai2012security,jain2011double}, poker games~\cite{waugh2009strategy} and search games~\cite{bosansky2012iterative} and it is widely used in various disciplines. 
Related to our research, DO is also used to alternately train the multiple generators and discriminators for the different architectures of GANs with best response oracles~\cite{aung2021gan}. We present the corresponding terminologies between GAN and game theory in Table~\ref{Comparison_Table}.

\begin{table*}[ht]
\centering
\caption{Comparison of Terminologies between Game Theory and GAN}
\begin{tabular}{ll}
\toprule\toprule
Game Theory terminology & GAN terminology \\ \midrule\midrule
Player                  & Generator/discriminator  \\ \midrule
Strategy                & The parameter setting of generator/discriminator, e.g., $\pi_g$ and $\pi_d$ \\\midrule
\multirow{2}{*}{Policy}               & The sequence of parameters (strategies) till epoch $t$, e.g., ($\pi^1_g, \pi^2_g, ..., \pi^t_g$) \\
& Note: Not used in this paper.\\\midrule
Game                    & The minmax game between generator and discriminator\\\midrule
\multirow{2}{*}{Meta-game/ meta-matrix}  & The minmax game between generator and discriminator with \\
& their respective set of strategies at epoch $t$ of DO framework                             \\\midrule
Meta-strategy           & The mixed NE strategy of generator/discriminator at epoch $t$  \\ \bottomrule\bottomrule                                    
\end{tabular}
\label{Comparison_Table}
\end{table*}

\section{Preliminaries}
In this section, we mathematically explain the preliminary works to effectively support our approach in addition to our previous work DO-GAN~\cite{aung2021gan}.
% In this section, we will provide a mathematical explanation of the preliminary works to effectively support our approach, building upon our previous work DO-GAN~\cite{aung2021gan}.
%including generative adversarial networks and game theory concepts such as normal-form game and double oracle algorithm.

\begin{figure*}[hb]
    \centering
    \includegraphics[width=0.8\linewidth]{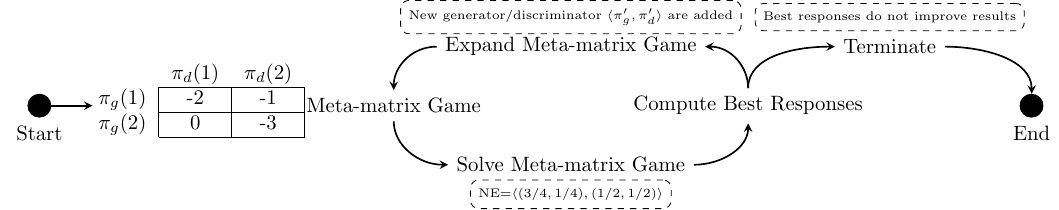}
    \caption{An illustration of the general framework for DONAS. Figure adapted from~\cite{lanctot2017unified}.}
    \label{fig:illustrative_do_gan}
\end{figure*}

\subsection{Adversarial Training}

Given a fixed classifier with
parameters $\theta$, a dataset $X$ (e.g., an image $x$ with true label $y$), and classification loss $\mathcal{L}$, a bounded non-targeted attack is done by
\begin{equation}
    \max_{\delta} \mathcal{L}(x + \delta, y, \theta), \delta\in S,
\end{equation}
where $S=\{ \delta ~\big \vert~ \left\Vert\delta \right\Vert_{p}\leq\epsilon\}$ is the space for the perturbation, $\left \Vert \cdot\right \Vert_{p}$ is the $p$-norm distance metric of adversarial perturbation and $\epsilon$ is the manipulation budget. We leverage Projected Gradient Descent (PGD) which uses the sign of the gradients and projects the perturbations into the set $S$, to construct an adversarial example for each iteration $t$ with step size $\epsilon$. 
\begin{equation}
    x^{t} = \Pi_{x+S} (x^{t-1} + \epsilon \cdot \mathrm{sgn}(\nabla_{x} \mathcal{L}(x^{t-1}, y, \theta))).
\label{advExampleConstruct}
\end{equation}
\noindent K-PGD~\cite{madry2017towards} is the iterative attack of the adversarial examples constructed in Eq. (\ref{advExampleConstruct}) with uniform random noise as initialization, i.e., $x^{0}=x + \zeta$ where $\zeta$ is from noise distribution. The strength of PGD attacks to generate adversarial examples depends on the number of iterations. It has an inner loop that constructs adversarial examples, while its outer loop updates the model using mini-batch stochastic gradient descent on the generated examples. However, it is generally slow. Thus, we adapt Free AT~\cite{shafahi2019adversarial} as the AT algorithm of our DO-Framework. It computes the ascent step by re-using the backward pass needed for the descent step. To allow for multiple adversarial updates to the same image without performing multiple backward passes, Free AT trains on the same mini-batch several times in a row.

\subsection{Double Oracle Framework for GANs}

DO-GAN~\cite{aung2021gan} translates GAN as a two-player zero-sum game between the generator player and the discriminator player. A two-player zero sum game is defined as a tuple $(\Pi, U) $ where $\Pi = (\Pi_{g}, \Pi_{d})$ is the set of strategies for generator player and discriminator player, $U : \Pi_{g} \times \Pi_{d} \rightarrow R$ is a utility payoff table for each joint policy played by both players. For zero-sum game, the gain of generator player $u(\pi_g, \pi_d)$ is equal to the loss of discriminator player $-u(\pi_g, \pi_d)$.

Then, at each iteration $t$, DO-GAN creates a restricted meta-matrix game with the newly trained generators $\pi_{g}'$ and discriminators $\pi_{d}'$ as strategies of the two players, calculates the payoffs, and expected utilities of mixed strategies to compute mixed NE of the restricted meta-matrix game $U^{t}$ between generator and discriminator players to get the probability distributions $(\sigma^{*}_{g}, \sigma^{*}_{d})$. Using the probability distribution, we compute the best responses and adds them into the restricted game for the next iteration. Figure~\ref{fig:illustrative_do_gan} presents an illustration of DO-GAN which is adapted as the general framework for extending the double oracle framework to NAS for GAN and AT. For the scalable implementation, DO-GAN uses pruning and continual learning techniques and removes the least-contributing strategies toward each player's winning to keep the number of generator and discriminator models from the best response at a size that the memory can support.

\section{DONAS for Game Theoretic Deep Learning}
In this section, we describe the double oracle frameworks with network architecture search for GAN and AT consisting of extending DO-GAN. Moreover, recent works propose evolutionary computation methods to neural architecture search methods for more compact and flexible architectures~\cite{xie2023point, zheng2021migo}. Inspired by the evolutionary methods, we propose Adversarial-Neural Architecture Search (DONAS-GAN) and Neural Architecture Search for Adversarial Training (DONAS-AT) to compute the mixed NE efficiently improving the generalizability and flexibility in model architecture. Over the different NAS methods for GAN and AT, the DO framework builds a restricted meta-matrix game between the two players and computes the mixed NE of the meta-matrix game, then iteratively adds more strategies of the two players (generators/discriminators or classifier/attacker) into the meta-matrix game until termination.

\subsection{General Framework for DONAS}
We have discussed the two prominent approaches to optimizing network architecture in adversarial machine learning, mainly, DO as the mixed-architecture approach, and Adversarial-NAS and AdvRush as the NAS approach for GANs and AT, respectively. DO finds the best response architecture by only updating weights for predefined architecture in the oracles, while NAS methods select only one final network from the search space and may miss the information trained by other top search architectures. To mitigate the issues in both methods, we propose the DONAS approach to combine them, allowing searched networks in a mixed-architecture framework.

We describe the general framework for our proposed DONAS. We adapt the double oracle framework, in which Neural Architecture Search is used as the best response oracle for the two players: generator-discriminator for GANs and classifier-attacker for AT. We search and sample the best response strategies for the two players with respective oracles. Then, we finetune to improve the quality of the image generation and robustness against adversarial attacks. To speed up the training, we introduce the terminating condition $\epsilon$.

\begin{algorithm}
\caption{DONAS: General Framework}
\small
\begin{algorithmic}[1]
\STATE $P_{1}, P_{2}$ = $\mathtt{InitializeModels()}$ \label{initializeG}
% \STATE $\mathcal{P}_{1}=\mathcal{P}_{2}=\emptyset$ \label{initializeGD}
% \STATE Initialize ; 
\STATE $\mathcal{P}_{1} \leftarrow \{P_{1}\}$, $\mathcal{P}_{2} \leftarrow \{P_{2}\}$, $\sigma^{1*}_{p1}=\sigma^{1*}_{p2}=[1]$\label{initializeD}
\FOR{epoch $t = 1,2, ...$}
\STATE $P_{1}= \mathtt{Player1\_NASSearch()}$
\label{searchD}
\STATE $\Pi_{1}$ = $\mathtt{SampleFromSupernet}(P_{1})$ \label{sample1}
\STATE $P_{2}= \mathtt{Player2\_NASSearch()}$
\label{searchGG}
\STATE $\Pi_{2}$ = $\mathtt{SampleFromSupernet}(P_{2})$ \label{sample2}
\STATE $\mathcal{P}_{1} \leftarrow \mathcal{P}_{1} \cup \Pi_{1}$ , $\mathcal{P}_{2} \leftarrow \mathcal{P}_{2} \cup \Pi_{2}$ \label{searchG}
\STATE $\mathtt{FineTuning()}$ \label{finetuneDONAS}
\STATE Augment $U^{t-1}$ with $P_{1}$ and $P_{2}$ to obtain $U^{t}$\label{payoffComputeNAS}
\STATE Compute mixed-NE $\langle\sigma^{t*}_{p1}, \sigma^{t*}_{p2}\rangle$ for $U^{t}$ \label{lpNAS}
\IF {$\mathtt{TerminationCheck}(U^{t}, \sigma^{t*}_{p1}, \sigma^{t*}_{p2})$} \label{terminationCheckNAS}
\STATE \textbf{break}
\ENDIF
%\STATE $\mathtt{PruneModels} (\mathcal{P}_{1}, U^{t},  \mathcal{P}_{2}, U^{t}, \sigma_{p1}^{*},\sigma_{p2}^{*})$ \label{Pruning}
\ENDFOR
\end{algorithmic}
\label{DONAS}
\end{algorithm}

Algorithm~\ref{DONAS} initializes the first pair of models and stores them in the two arrays $\mathcal{P}_{1}$ and $\mathcal{P}_{2}$. Then, we can randomly sample pure strategies to augment the meta-game and compute the distribution of the player strategies. We initialize the meta-strategies $\sigma^{1*}_{p1}=[1]$ and $\sigma^{1*}_{p2}=[1]$ since only one pair of player strategies is available (line~\ref{initializeG}-\ref{initializeD}). For each epoch, we search the new strategy for each player (line~\ref{searchD},~\ref{searchGG}), obtaining the supernets $P_{1}$ and $P_{2}$. Next, we sample the networks to derive the architecture from supernet (line~\ref{sample1},~\ref{sample2}). We then finetune the selected models (line~\ref{finetuneDONAS}). In DONAS for AT, we do not have sampling and finetuning for the best response oracle for attacker player since we do not need to search architecture for the attacker player. Next, we generate the meta-game $U^{t}$ and compute mixed NE with linear programming proposed in~\cite{aung2021gan} to obtain $\sigma^{*}_{p1}$ and $\sigma^{*}_{p2}$ (line~\ref{payoffComputeNAS}-\ref{lpNAS}).  The algorithm terminates if the terminating condition $\epsilon$ is satisfied (line~\ref{terminationCheckNAS}) which we follow the procedures of the DO-GAN/P~\cite{aung2021gan}.
\subsection{DONAS for GAN}
In this section, we use DONAS for GANs to allow the use of multiple generators and discriminators. We let $\mathcal{P}_{1}=\mathcal{G}$ and $\mathcal{P}_{2}=\mathcal{D}$ as the two-player zero-sum game between the generator and discriminator with their architectures as $\alpha$ and $\beta$ respectively. Theoretically, using multiple generators and discriminators with mixed strategies covers multiple modes to produce better generated images~\cite{arora2017generalization,hoang2018mgan, hsieh2018finding}. 

\begin{algorithm}
\caption{DONAS for GANs}
\begin{algorithmic}[1]
\small
\ENSURE $\mathtt{InitializeModels()}$
\STATE Initialize the generator $G$ and discriminator $D$
\label{searchGNAS}
% \STATE D= $\mathtt{InitializeDiscriminator()}$ 
\STATE $\mathtt{FineTune(G)}$ \label{searchDNAS}
\ENSURE $\mathtt{GeneratorOracle()}$
\STATE $G$ = initialize()
\FOR{$s$ steps}
\STATE Sample mini-batch of $2m$ noise samples. \label{sampleG}
\STATE Update the architecture of generator: \\
$\nabla_{\alpha} \frac{1}{m}\sum \limits^{m}_{i=1} [\sum\limits_{j=1}^{|\mathcal{D}|} \sigma_{d}^{j*} \cdot \log(1-D_{j}(G(z^{i})))]$ \label{updatearchG}
\STATE Update the weights of generator:\\
$\nabla_{W_{G}} \frac{1}{m}\sum\limits^{2m}_{i>m} [\sum\limits_{j=1}^{|\mathcal{D}|} \sigma_{d}^{j*}\cdot\log(1-D_{j}(G(z^{i})))]$ \label{updateweightsG}
\ENDFOR
\ENSURE $\mathtt{DiscriminatorOracle()}$
\STATE $D$ = initialize()
\FOR{$s$ steps} 
\STATE Sample $2m$ samples for noise and real data\label{samplenoise}
% \STATE Sample mini-batch of $2m$ real data examples\label{samplereal}
\STATE Update the architecture of discriminator: \\
$\nabla_{\beta} \frac{1}{m}\sum^{m}_{i=1}[\log x^{i}+ \sum_{j=1}^{|\mathcal{G}|} \sigma_{g}^{j*}\cdot\log(1-D(G_{j}(z^{i})))]$ \label{updatearch}
\STATE Update the weights of discriminator:\\
$\nabla_{W_{D}} \frac{1}{m}\sum^{2m}_{i>m} [\log x^{i}+ \sum_{j=1}^{|\mathcal{G}|} \sigma_{g}^{j*} \cdot \log(1-D(G_{j}(z^{i})))]$ \label{updateweights}
\ENDFOR
\ENSURE $\mathtt{SampleFromSupernet}(\alpha)$
\STATE Sample top $k$ subnetworks $\bar{\alpha}$ according to $\alpha$
\label{samplesupernet}
\STATE $\Pi = \text{argmin}_{\Pi \in \bar{\alpha}} \text{ adv\_loss()}$
\label{samplesupernet2}
\ENSURE $\mathtt{SequentialFineTune()}$
\FOR{$r=1,2, ...$} \label{finetuneGDstart}
\FOR{$i=1, \dots,|\mathcal{G}|$}
\STATE \small $\max_{\theta_{g_{i}}} \mathop{\mathbb{E}}_{z} \big[ \sum_{j=1}^{|\mathcal{D}|} \sigma^{j*}_{d}\cdot D_{j}(G_{i}(z)) \cdot \Theta \big]$ \label{finetuneG}
\ENDFOR 
\FOR{$j=1, \dots,|\mathcal{D}|$}
\STATE $\max_{\theta_{d}} \mathop{\mathbb{E}}_{x} [ \log D_{j}(x)] + \sum_{k=1}^{K} \mathop{\mathbb{E}}_{z}[\log(1 - D_{j}(G_{k}(z)))]$ \label{finetuneGDend}
\ENDFOR 
\STATE Augment $U^{t}$ and find mixed-NE $(\sigma_{g}^{*},\sigma_{d}^{*})$ \label{augmentU}
% \STATE Compute mixed-NE $(\sigma_{g}^{*},\sigma_{d}^{*})$ for $U^{t}$ \label{solveU}
\ENDFOR
\end{algorithmic}
\label{AdvNASGAN}
\end{algorithm}
Algorithm~\ref{AdvNASGAN} describes specific changes to adapt DONAS for GANs. We initialize the first generator-discriminator pair $G,D$ and finetune the models in $\mathtt{InitializeModels()}$ (line~\ref{searchGNAS}-\ref{searchDNAS}). The two-player oracles are shown in $\mathtt{{GeneratorOracle}()}$ and $\mathtt{DiscriminatorOracle()}$ which describe the best response oracles in the DO-framework of a two-player min-max game with networks $(\alpha$ and $\beta)$ where the weight of each network must be the best response to the other player. 

\noindent \textbf{Adapting Adversarial-NAS.} GAN optimization process in Adversarial-NAS is defined as a two-player min-max game with value function $V (\alpha, \beta)$ where the weight of each network must
be the best response~\cite{gao2020adversarialnas}. The min-max game is $\min \nolimits_{\alpha} \max \nolimits_{\beta} V (\alpha, \beta)$ as:
% \begin{small}
\begin{align}
V (\alpha, \beta)= \mathbb{E}_{x\sim p_{data}}(x) [\log D(x \vert \beta, W^{*}_{D}(\beta)] + \notag\\ \mathbb{E}_{z \sim p_{z}(z)} [\log(1- D(G(z \vert \alpha, W^{*}_{G} (\alpha)) \vert \beta, W^{*}_{D}(\beta))],
\end{align}
% \end{small}
\normalsize
\noindent where the two weights ${W^{*}_{G}(\alpha), W^{*}_{D}(\beta)}$ for any architecture pair $(\alpha, \beta)$ can be obtained through another min-max game between $W_{G}$ and $W_{D}$, i.e., $\min \nolimits_{W_G(\alpha)} \max \nolimits_{W_D(\beta)} V (W_{G(\alpha)}, W_{D(\beta)})$ as:
% \begin{small}
\begin{align}
    &V (W_{G(\alpha)}, W_{D(\beta)}) = \mathbb{E}_{x\sim p_{data}}(x) [\log D(x \vert\beta, W_{D(\beta)}] + \notag \\ & \mathbb{E}_{z \sim p_{z}(z)} [\log(1- D(G(z \vert \alpha, W_{G(\alpha)})  \vert\beta, W_{D(\beta)}))].
\end{align}
% \end{small}
\normalsize
\noindent The weights of generator and discriminator w.r.t. the architecture pair $(\alpha, \beta)$, $\{W^{*}_{G(\alpha)}, W^{*}_{D(\beta)}\}$, can be obtained by a single step of adversarial training as vanilla GANs~\cite{shafahi2019adversarial,gao2020adversarialnas}. The optimal architectures or weights in each iteration can be achieved by ascending or descending the corresponding stochastic gradient by updating in the order of architecture followed by the weights. 

\noindent \textbf{Generator Oracle.} We initialize the generator and sample the noise samples (line~\ref{sampleG}). Then, we search and update the architecture of the generator by descending its stochastic gradient using the probability distribution $\sigma^{*}_{d}$ for the pure strategies of the discriminator (line~\ref{updatearchG}). Finally, we update the weight of the generator by the stochastic gradient descent while considering $\sigma^{*}_{d}$ (line~\ref{updateweightsG}).

\noindent \textbf{Discriminator Oracle.} Similarly, we initialize the discriminator, sample noise samples and real-data examples (line~\ref{samplenoise}). Then, we update the architecture (line~\ref{updatearch}) and weights (line~\ref{updateweights}) by stochastic gradient ascent with the probability distribution $\sigma^{*}_{g}$ for the pure strategies of the generator.

\noindent \textbf{Sampling Architecture from Supernet.}
After the search, we derive the architecture from the resulting supernet (line~\ref{samplesupernet}-~\ref{samplesupernet2}). We sample top $k$ networks selecting maximum values of $\alpha$ and select the networks that give the minimum adversarial loss against the other player.

\noindent \textbf{Sequential Finetuning.} After the two oracles provide the new strategies for the two players (new generator and discriminator), we sequentially finetune them by Nash distributions $\sigma^{0*}_{g}$ and $\sigma^{0*}_{d}$ that we obtained from solving the meta-game with linear programming (line~\ref{finetuneGDstart}-\ref{finetuneGDend}).
To finetune multiple generators altogether, we update the objective function of the generator in training with $\mathbb{E}_{(z\sim p_{z}(z))} [D(G_{i}(z, \theta_{gi}))]$ and the sequential generators with harmonic mean \cite{varshney2020stm}. Hence, the generator update is:
\begin{equation}
    \max_{\theta g_{i}} \mathbb{E}_{z\sim p_{z}(z)} [D(G_{i}(z, \theta_{gi}))\cdot\Phi],
\label{Harmonic}
\end{equation}
where $\Phi$ is the harmonic mean of the remaining generators in the array: $\Phi=HM((1 - D(G_{i-1}(z, \theta_{g_{i-1}})), \dots,(1 - D(G_{1}(z, \theta_{g_{1}})))]$. 

This ensures that the current generator focuses on the data samples from the modes ignored by the previous generators. Arithmetic means and geometric means cannot generalize the ignored data samples well in the case of sampled generators where one of the generators is performing much better.
The Harmonic Mean makes sure the currently trained generator focuses on the ignored data samples. The details of sequential finetuning with HM are mentioned in Algorithm~\ref{finetune}.

\begin{algorithm}[ht]
\caption{Fintuning $K$ Generators}
\begin{algorithmic}[1]
\FOR{iteration 1,2, ...}
\FOR{$i=1, \dots,K$}
\STATE Update the generators using Harmonic Mean: \\$\max_{\theta_{g_{i}}} \mathbb{E}_{z\sim p_{z}(z)} [D(G_{i}(z, \theta_{gi})) 
    \times HM((1 - D(G_{i-1}(z, \theta_{g_{i-1}}))), 
    \dots,(1 - D(G_{1}(z, \theta_{g_{1}}))))]$
\ENDFOR
\STATE Update the discriminator:\\ 
$\max_{\theta_{d}} \mathbb{E}_{x\sim P_{data}} [\log D(x, \theta_{d})] + \sum_{k=1}^{K} \mathbb{E}_{z-p_{z}(z)}[log(1 - D(G_{k}(z, \theta_{g_k}), \theta_{d}))] $ 
\ENDFOR
\end{algorithmic}
\label{finetune}
\end{algorithm}

The sequential finetuning of $K$ generators by Harmonic Mean is shown in Algorithm \ref{finetune}. For every epoch, each generator is updated according to Eq. (\ref{Harmonic}) (line 3) followed by the discriminator update where the fake data from the generator is the combination of all $K$ generators (line 5). After the training has finished, we augment the meta-game and solve it to obtain the distribution of $K$ generators to be used in the next iteration. However, HM only tells the ignored samples and cannot give a clear picture of which samples are being captured more than the others at each epoch of the finetuning. 

Using the meta-game, which only takes linear complexity to solve, gives Nash distribution of the generators, and hence, the update of the objective function can have a better picture to focus on the less-captured data samples. This method is more effective than focusing heavily on the ignored data samples.
Algorithm~\ref{AdvNASGAN}: $\mathtt{SequentialFineTune()}$ describes the finetuning of multiple generators and discriminators with Nash. Using a similar equation as Eq.~(\ref{Harmonic}) and $\Theta = \sigma^{i-1*}_{g}(1 - D_{j}(G_{i-1}(z, \theta_{g_{i-1}}))) \times \dots \times \sigma^{1*}_{g}(1 - D_{j}(G_{1}(z, \theta_{g_{1}})))$,  we finetune each generator (line~\ref{finetuneG}) and discriminator (line~\ref{finetuneGDend}) using Nash distribution. Instead of using HM, we use the Nash distribution from generating the augmented meta-game and compute mixed NE by linear programming (line 24).

\iffalse 
\begin{algorithm}[ht]
\caption{DO-Adverarial Training}
\begin{algorithmic}[1]
\STATE $\mathcal{C}=\emptyset, \mathcal{a}=\emptyset$, \label{initializeModels}
\STATE $\theta = \mathtt{AdvRushSearch()}$ \label{classifier}
\STATE $\delta$ = 0 \label{attacker}
\STATE Initialize $\sigma^{1*}_c =[1]$ and $\sigma^{1*}_p=[1]$; 
\STATE $\mathcal{C} \leftarrow \mathcal{C} \cup \theta$, $\mathcal{a} \leftarrow \mathcal{a} \cup \delta$; \label{initializeP}
\FOR{epoch $t = 1,2, ...$}
\STATE $\delta = \mathtt{AttackerOracle()}$ \label{searchA}
\STATE $\mathcal{a} \leftarrow \mathcal{a} \cup \delta$ 
\STATE $\theta = \mathtt{ClassifierOracle()}$
\STATE $\mathcal{C} \leftarrow \mathcal{C} \cup \theta$ \label{searchC}
\STATE Augment the meta game $U^{t}$ with $adv\_loss$ \label{augmentUAT}
\STATE $\sigma_{c}^{*},\sigma_{a}^{*}  =$ solve $U^{t}$ \label{solveUAT}
\IF {\texttt{TerminationCheck}($U^{t}, \sigma^{t*}_c, \sigma^{t*}_p$)} \label{terminationCheckAT}
\STATE break
\ENDIF
\STATE $\mathtt{PruneModels} (\mathcal{a}, U^{t},  \mathcal{C}, U^{t}, \sigma_{p}^{*},\sigma_{c}^{*})$ \label{PruningAT}
\ENDFOR
\end{algorithmic}
\label{DOAdvRush}
\end{algorithm}
\fi

\begin{algorithm}
\caption{DONAS for AT}
\begin{algorithmic}[1]
\REQUIRE Samples $X$, learning rate $\gamma$, the number of hopsteps $m$
\ENSURE $\mathtt{InitializeModels()}$
\STATE $\theta = \mathtt{AdvRushSearch()}$ \label{classifier}
\STATE $\delta$ = 0 \label{attacker}
\ENSURE $\mathtt{AttackerOracle()}$
\STATE Let $\delta=0$
\FOR{step $s = 1,2, ..., \frac{N}{m}$}
\FOR{mini-batch $B \subset X$}
\FOR{$i=1,2, \dots, m$}
\STATE $g_{adv} = \nabla_{x, \theta\sim (C,\sigma^{*}_{c})} l(x+\delta, y, \theta), x,y \in B$ \label{calculateAdvGrad}
\STATE Update $\delta$ \label{updatedelta}\\
$\delta \leftarrow \mathtt{Clip}(\delta + \epsilon \cdot sign(g_{adv}), -\epsilon, \epsilon)$
\ENDFOR
\ENDFOR
\ENDFOR
\ENSURE $\mathtt{Classifier Oracle()}$
\STATE Initialize architecture and weight $(w_{0},\alpha_{0})$
\FOR{iteration $i=1, 2, ...$}
\STATE Update weights by descending its stochastic gradient:\\ $\nabla_{W} L_{train}(w_{i-1}, \alpha_{i-1})$
\STATE  Update the architecture, i.e., $\alpha$: \\
$
\begin{cases}
\nabla_{\alpha} [L_{val}(w_{i}, \alpha_{i-1})], \quad &\text{if $t<\phi$}\\
\nabla_{\alpha} [L_{val}(w_{i}, \alpha_{i-1}) + \gamma L_{\lambda}], &\text{else}
\end{cases}
$\label{advsearch}
\ENDFOR
\ENSURE $\mathtt{SampleFromSupernet}(\theta)$
\STATE Derive $\theta$ through discritization steps from DARTS
\ENSURE $\mathtt{FineTune()}$
\FOR{steps $s = 1,2, ...,\frac{N}{m}$} \label{finetunestart}
\FOR{mini-batch $B \subset X$}
\FOR{$i=1,2, \dots, m$}
\STATE $g_{\theta} \leftarrow \mathbb{E}_{(x,y)\in B} [\nabla_{\theta, \delta \sim (\mathcal{A}, \sigma^{*}_{a}}) l(x + \delta, y, \theta)]$ \label{calculateGradient}
\STATE $\theta \leftarrow \theta - \gamma \cdot  g_{\theta}$ \label{updatetheta}
\ENDFOR
\ENDFOR
\ENDFOR
\end{algorithmic}
\label{DONASAT}
\end{algorithm}

\subsection{DONAS for AT}

We also propose DONAS for AT where we have the two players as the training classifier $\mathcal{P}_{1}$= $\mathcal{C}$ and the adversarial $\mathcal{P}_{2}$= $\mathcal{A}$. We search the classifier in $\mathtt{ClassifierOracle()}$ against the perturbed datasets from $\mathtt{AttackerOracle()}$. 
Given $L_{val}$ and $L_{train}$ as loss functions on the validation and training sets respectively, the search process of NAS is defined as a bi-level optimization problem:
% \begin{small}
\begin{align}
\label{bilevelsearch}
    &\min\nolimits_{\theta}  L_{val}(w^{*}(\theta), \theta)
    \notag \\  \text{s.t.} &~ w^{*}(\theta) = \text{argmin}_{w} L_{train}(w, \theta).
\end{align}
% \end{small}
\normalsize
\noindent For NAS in the classification task,
the expensive inner optimization in Eq.~(\ref{bilevelsearch}) is normally approximated by one step training~\cite{liu2018darts} as:
$\nabla_{\theta} L_{val} (w^{*}(\theta), \theta) \approx \nabla_{\theta} L_{val}(w- \xi \nabla_{w} L_{train}(w, \theta), \theta)$ where $w$ denotes the current weights and $\xi$ is the learning rate for a step
of inner optimization. 

\begin{table*}[!ht]
\centering
\setlength\tabcolsep{4pt}
\caption{Generative results for DONAS-GAN.}
\begin{tabular}{c|c|c|cc|ccccc}
\toprule
\multicolumn{3}{c|}{\multirow{2}{*}{Methods}}  & \multirow{2}{*}{AutoGAN}     & \multirow{2}{*}{AdvNAS} & \multicolumn{5}{c}{DONAS-GAN}\\
\cmidrule{6-10}
\multicolumn{3}{c|}{}& &   & {Ind.} & {HM} & {Nash.} & {Iter.} & Full\\ \midrule
\multirow{4}{*}{CIFAR-10}      & \multirow{2}{*}{Search}   & 5  &    \multirow{2}{*}{-}      & \multirow{2}{*}{16.35} & 14.22   & 14.22 & 14.22    & 14.43    &   - \\   &      & 10 &   &   & 14.04   & 14.04 & 14.04    & 13.89    &   - \\\cmidrule{2-10}
   & \multirow{2}{*}{Finetune} & 5  & \multirow{2}{*}{12.42} & \multirow{2}{*}{\underline{10.87}}    & 10.62   & 9.27  & 9.06     & 9.12     & \textbf{8.93}     \\
   
   &      & 10 &   &    & 10.33   & 9.32  & 9.19     & 9.10      & \textbf{8.93}     \\ \midrule
\multirow{4}{*}{STL-10} & \multirow{2}{*}{Search}& 5  &    \multirow{2}{*}{-}  &\multirow{2}{*}{32.48}    & 29.16   & 29.16 & 29.16    & 30.02    &  -  \\
   &      & 10 &   &   & 28.95   & 28.95 & 28.95    & 28.99    &     -      \\\cmidrule{2-10}
   & \multirow{2}{*}{Finetune}    & 5  & \multirow{2}{*}{31.01} & \multirow{2}{*}{\underline{26.98}} & 28.88   & 26.44 & 26.13    & 26.15    & \textbf{25.31}     \\
   &      & 10 &   &   & 28.72   & 26.60  & 26.17    & 25.82    & \textbf{24.75}     \\ \midrule
\multirow{4}{*}{\makecell[c]{Tiny \\ ImageNet}} & \multirow{2}{*}{Search}& 5  &      \multirow{2}{*}{-}    &     \multirow{2}{*}{17.99} & 17.53 &    17.53   & 17.53  &   -&   - \\
&& 10 &   &      &     16.91      &    16.91     & 16.91    &    -      &   - \\\cmidrule{2-10}
  & \multirow{2}{*}{Finetune}    & 5  & \multirow{2}{*}{16.21} & \multirow{2}{*}{\underline{15.10}}  &  15.21&  13.85     &  12.36 &    -      & \textbf{11.18}     \\
   &      & 10 &   &   &   14.60      &  13.28     & 11.94  &    -      & \textbf{10.42}  \\ \bottomrule
\end{tabular}
\label{maintable}
\end{table*}

\noindent \textbf{Adapting AdvRush.}
Algorithm~\ref{DONASAT} describes the specific changes to adapt DONAS for AT. To obtain the first pair of classifier $\theta$ and perturbed dataset $\delta$, we perform neural architecture search for $\theta$ by AdvRush~\cite{mok2021advrush} and set $\delta=0$ in $\mathtt{InitializeModels()}$, then store the parameters in the two arrays $\mathcal{C}$ and $\mathcal{A}$ (line~\ref{classifier}-\ref{attacker}). For each epoch, we search the new classifier and generate a new perturbed dataset with $\mathtt{ClassierOracle()}$ and $\mathtt{AttackerOracle()}$ using the meta-strategies $\sigma^{*}_c$ and $\sigma^{*}_p$. We refer to the Free adversarial training (Free AT) algorithm~\cite{shafahi2019adversarial}, with comparable robustness to the traditional 7-step PGD~\cite{madry2017towards} with significantly faster training. We then convert it to the DO-Framework with the two oracles to allow multiple adversarial updates to be made to the same images without multiple backward passes by training the same mini-batch for $m$ times.

\noindent \textbf{Attacker Oracle.} For each iteration, we calculate the adversarial gradient using $\theta \in \mathcal{C}$ sampled with the probability distribution of the classifier's strategies from the meta-game $\sigma_{c}^{*}$ to update $\delta$ (line~\ref{calculateAdvGrad}-\ref{updatedelta}).

\noindent \textbf{Classifier Oracle.} We adapt AdvRush to search for the architecture and update the weights. According to the ablation study by~\cite{mok2021advrush}, we set $\gamma=0.01$ and AdvRush loss term $L_{\lambda}$ is introduced at $\phi=50$ which we set as iteration number for warm-up process. (line ~\ref{advsearch}).

\noindent \textbf{Sampling Architecture from Supernet.} After the search and update of the classifier's architecture and weights, we derive the final architecture (line~\ref{samplesupernet}) by running the discretization procedure of DARTS~\cite{liu2018darts}. 

\noindent \textbf{Finetuning.} 
We adversarially train the classifier against the attacker. We update $\theta$ with stochastic gradient descent (line ~\ref{calculateGradient}-\ref{updatetheta}) with $\delta$ sampled from $\mathcal{A}$ with $\sigma^{*}_{a}$. We consecutively train each mini-batch for $m$ times. 

Finally, we augment the meta-game by calculating the cross-entropy loss and solve it to obtain the distribution. DONAS-AT ends when terminating criteria is satisfied meaning both oracles have searched the best response strategies.

\section{Experiments}
We conduct our experiments on a machine with Xeon(R) CPU E5-2683 v3@2.00GHz and $4 \times$ Tesla v100-PCIE-16GB running Ubuntu operating system. We extend to NAS architectures such as Adversarial-NAS, AdvRush and evaluate against SOTA architectures such as AutoGAN, ResNet-18 and RobNet. We adopt the parameter settings and criterion of the baselines as published.  We set $s=10$ unless mentioned otherwise. We compute the mixed NE of the meta-game with Nashpy.

\subsection{Experiments on DONAS for GAN}

We first conduct an experiment to evaluate DONAS-GAN on the datasets: CIFAR-10~\cite{cifarten}, STL-10~\cite{coates2011analysis} and TinyImageNet~\cite{Le2015TinyIV} and evaluate the quantitative performance with FID score against the baselines AutoGAN~\cite{gong2019autogan} and Adversarial NAS~\cite{gao2020adversarialnas} to compare with our DONAS-GAN variants where we investigated ablation studies with different approaches varying the number of generators and discriminators.

\noindent \textbf{Variants of DONAS-GAN.}
Initially, we reproduce the Adversarial-NAS search which searches 1 generator followed by sampling multiple top networks from the supernet after the search. We run the experiments for DONAS-GAN/Ind., DONAS-GAN/HM and DONAS-GAN/Nash with $K$, where $K$ is the number of generators, to select a sample of multiple $K$ generators instead of 1 maximum and finetune the generators. We set the number of discriminators as 1. DONAS-GAN/Ind. uses independent finetuning where all $K$ generators are trained independently whereas DONAS-GAN/HM and DONAS-GAN/Nash use sequentially finetuning where the sampled $K$ generators are finetuned by Harmonic Mean (HM) and Nash. Next, we search $K$ generators iteratively in DONAS-GAN/Iter. and sample the architecture with the maximum $\alpha$ instead of a one-time search. To avoid having the searching architecture for the same batches, we shuffle the train data for each iteration.  After the iterative process, we finetune the resulting $K$ generators by Nash. Finally, we perform experiments with the full-version of DONAS-GAN with the pruning limit of $K$ for both generators and discriminators. We set $K$ to prune the generator and discriminator arrays $\mathcal{G}$ and $\mathcal{D}$, and the terminating condition $\epsilon$ is now set as $5e^{-3}$.

\begin{figure}
    \centering
    \includegraphics[width=0.7\linewidth]{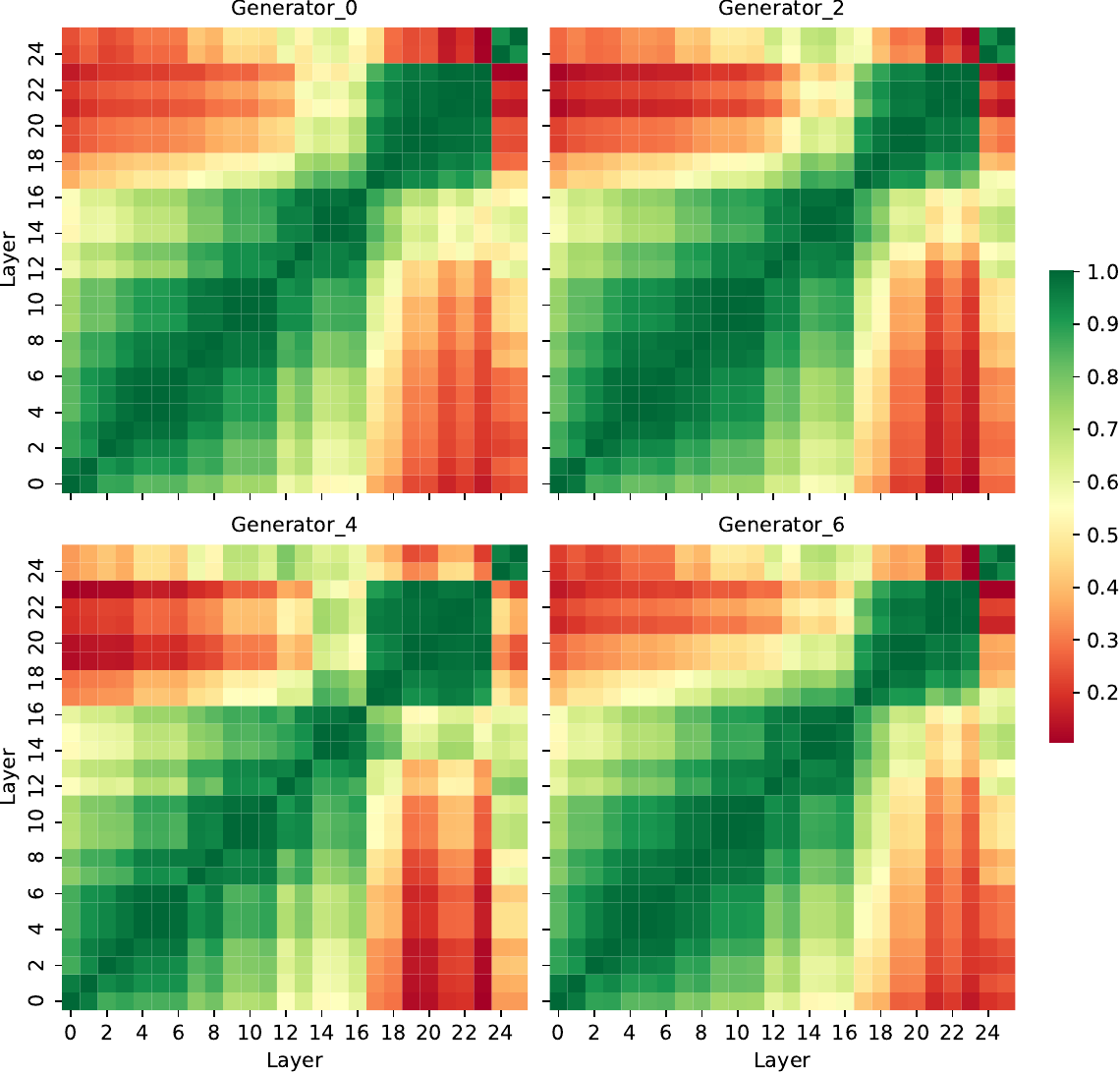}
    \caption{Linear CKA between layers of the individual searched networks of DONAS-GAN trained on TinyImageNet dataset.}
    \label{fig:cka_generator_self}
\end{figure}

\noindent \textbf{Experiments Results.} Table~\ref{maintable} shows the experiment results with various methods and datasets investigated on DONAS-GAN. We initially set up the experiment for the CIFAR-10 dataset and recorded the FID scores: AutoGAN as $12.42$, Adversarial-NAS as $16.35$ for the searched architecture and $10.87$ after finetuning. For the variants of DONAS-GAN, we recorded $10.62, 9.27, 9.06, 9.12, \mathbf{8.926} (K=5)$ and $10.33, 9.32, 9.19, 9.10, \mathbf{8.932} (K=10)$  for DONAS-GAN Ind., HM, Nash, Iter and full-version of DONAS-GAN with $K$ generators/discriminators. The results indicate that our variants bring significant improvements to the results. While DONAS-GAN Iter. suffers from long finetuning time as there is no terminating condition recording $60.88$ GPU Hours for $K=5$ and $109.52$ GPU Hours for $K=10$, DONAS-GAN shows satisfying results terminating within a training time of $54.19$ GPU Hours for the best run.

Similar promising trends are observed in STL-10 datasets recording $31.01$ and $26.98$ for AutoGAN and Adversarial-NAS as baselines. We recorded $28.88, 26.44, 26.13, 26.15, \mathbf{25.31} (K=5)$ and $28.72, 26.6,$ $26.17, 25.82, \mathbf{24.75} (K=10)$ for DONAS-GAN/Ind., HM, Nash, Iter and full-version of DONAS-GAN with $K$ generators/discriminators respectively. We also compared DONAS-GAN against baselines for TinyImageNet dataset and recorded $16.21$ for AutoGAN, $15.10$ for Adversarial-NAS and $11.18 (K=5), 10.42 (K=10)$ for DONAS-GAN showing that DONAS-GAN can capture the diversity of data examples better. Qualitative examples in Appendix~\ref{appendixCDONAS} also indicate the realistic image generation and the ability to capture the diversity of the classes without mode-collapse. 

\noindent \textbf{Internal Representation of Networks.} To analyze how the representation of the networks evolve and differ from each other, we use the centered kernel alignment (CKA)~\cite{kornblith2019similarity} to measure the similarity of the representation between layers and networks of DONAS-GAN trained on TinyImageNet. 
%{moved to appendix due to space issue}
%proposed by Kornblith et al to measure the similarity of the representation between layers and networks of DONAS-GAN trained on TinyImageNet. CKA is a commonly used metric for representation similarity, which features in higher accuracy comparing with other similarity indexes. 
In Fig.~\ref{fig:cka_generator_self}, we visualize the CKA values of layers in the same network of DONAS-GAN as heatmaps. We can see that there is a considerable similarity between the hidden layers as represented by the green area but the similarity goes down at the later layers. The trend indicates that our generation quality progressively improves with depth.

\begin{table*}[!ht]
\centering
\setlength\tabcolsep{4pt}
\caption{Robust accuracy under FGSM and PGD attacks.}
\begin{tabular}{c|l|c|ccc|c|c}
\toprule
\multicolumn{2}{c|}{\multirow{2}{*}{Methods}}              & \multirow{2}{*}{ResNet-18} & \multicolumn{3}{c|}{RobNet} & \multirow{2}{*}{AdvRush} & \multirow{2}{*}{DONAS-AT} \\ \cmidrule{4-6}
\multicolumn{2}{c|}{} & & Large & LargeV2 & Free & \\\midrule
\multirow{3}{*}{CIFAR-10}     & FGSM    & 49.81\%   & 54.98\%      & 57.18\%        & 59.22\%     & 60.87\% & \textbf{62.31\%}  \\ 
  & PGD-20  & 45.86\%   & 49.44\%      & 50.53\%        & 52.57\%     & 53.07\% & \textbf{59.57\%}  \\
  & PGD-100 & 45.10\%   & 49.24\%      & 50.26\%        & 51.14\%     & 52.80\% & \textbf{57.20\%}  \\ \midrule
\multirow{3}{*}{SVHN}         & FGSM    & 88.73\%   & 93.01\%      &     -           & 93.04\%     & 94.95\% & \textbf{95.91\%}  \\
  & PGD-20  & 69.51\%   & 86.52\%      &      -          & 88.50\%     & 91.14\% & \textbf{91.40\%}  \\
  & PGD-100 & 46.08\%   & 78.16\%      &       -         & 78.11\%     & 90.48\% & \textbf{91.83\%} \\ \midrule
\multirow{3}{*}{\makecell[c]{Tiny \\ ImageNet}} & FGSM    & 16.08\%   & 22.16\%      &     -           & 23.11\%     & 25.20\% & \textbf{28.11\%}  \\
                         & PGD-20  & 13.94\%   & 20.40\%      &         -       & 21.05\%     & 23.58\% & \textbf{26.30\%}  \\
  & PGD-100 & 13.96\%   & 19.90\%      &     -           & 20.87\%     & 22.93\% & \textbf{24.10\%}  \\ \bottomrule
\end{tabular}
\begin{tablenotes}
    \centering
      \footnotesize
      \item Note: For ResNet-18 and RobNets, we presented the results from the papers~\cite{mok2021advrush,guo2020meets}.
    \end{tablenotes}
\label{ATresultTable}
\end{table*}

\begin{figure}
    \centering
    \includegraphics[width=0.7\linewidth]{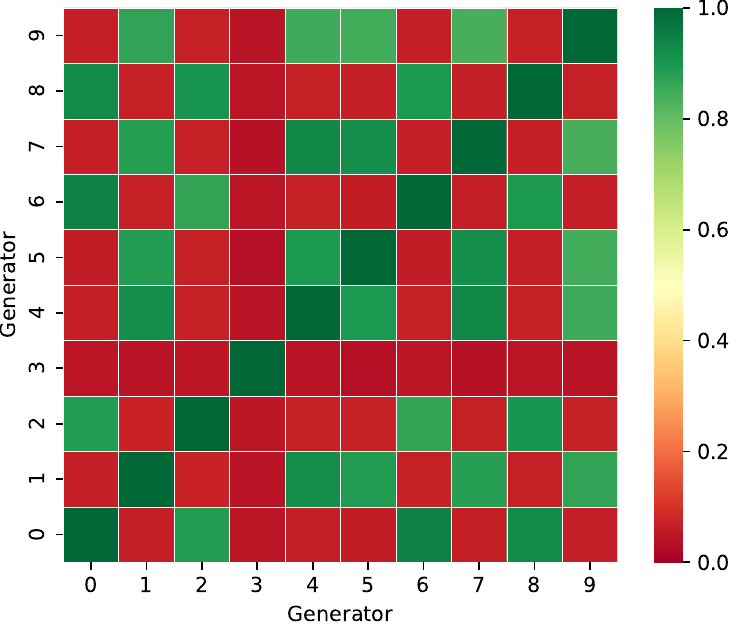}
    \caption{Linear CKA averaging between the same layers of DONAS-GAN searched networks trained on TinyImageNet. Generators 0, 1 and 3 have different architectures.}
    \label{fig:cka_generator_mutual_line}
\end{figure} 
In Fig.~\ref{fig:cka_generator_mutual_line}, we visualize the similarity across the models. The result indicates that the search process of DONAS-GAN can obtain diverse architectures of generators (e.g., Generators 0, 1, and 3) to be used as mixed-architecture mitigating the mode-collapse problem. We present the full network similarity analysis in Appendix~\ref{appendixEDONAS}.

%We also observe some difference in the searched networks indicating that DONAS-GAN also can obtain diverse generator networks as shown in Fig.~\ref{fig:cka_generator_self_2}.

\iffalse
\begin{figure}
    \centering
    \includegraphics[width=\linewidth]{cka_figs/cka_generator_self_2.pdf}
    \caption{Caption}
    \label{fig:cka_generator_self_2}
\end{figure}

\begin{figure}
    \centering
    \includegraphics[width=\linewidth]{CIFAR-10.png}
    \caption{The CIFAR-10 images randomly generated from DONAS for GAN using $K=10$.}
    \label{fig:cifar10}
\end{figure}
\fi

\subsection{Experiments on DONAS for AT}

Finally, we evaluate the robustness of the architectures searched and adversarially trained by DONAS-AT on CIFAR-10 using FGSM and PGD white-box-attacks~\cite{madry2017towards}. In particular, we evaluate our proposed DONAS-AT with ResNet-18~\cite{he2016deep}, variations of RobNet~\cite{guo2020meets} and finally, AdvRush~\cite{mok2021advrush} on CIFAR-10, TinyImageNet and SVHN datasets. According to~\cite{guo2020meets}, the model's robustness is better as we increase the computational budget and thus, we compare with RobNet-Large and RobNet-LargeV2 which have comparable network parameters (number of channels and cells) to WideResNet. Moreover, we also select RobNet-Free which relaxes the cell-based constraint.
After searching the classifier and training in DONAS-AT, we attack the model using PGD attacks where perturbation data is added to input for $T$ iterations ($PGD^{T}$). We then use this perturbed input to the model for classification and calculate cross-entropy loss against the target label as criterion. We test the classifiers using $PGD^{20}$ and increase the iterations to 100, i.e., $PGD^{100}$, to introduce stronger attacks.

Evaluation results are shown in Table~\ref{ATresultTable}. We observe promising results of DONAS-AT with $62.31\%$ accuracy for FGSM attacks on the CIFAR-10 dataset surpassing the baseline methods such as ResNet-18, RobNet-Large, RobNet-LargeV2, RobNet-Free and AdvRush that record $49.8\%, 54.98\%, 57.18\%, 59.22\%, 60.87\%$. We also evaluated with PGD white-box attacks $PGD^{20}$ and $PGD^{100}$. We obtain $59.57\%$ accuracy on $PGD^{20}$ white-box attacks and  $57.20\%$ for $PGD^{100}$ observing similar trends which demonstrates the stronger robustness of our approach.

\section{Conclusion}
In this paper, we propose the double oracle framework solution to NAS for game theoretic deep learning models such as GAN as well as AT. The double oracle framework starts with a restricted game and incrementally adds the best responses of the player oracles (either for generator/discriminator or classifier/attacker) as the players' strategies. We then compute the mixed NE to get the players' meta-strategies by using a linear program. We leverage two NAS algorithms Adversarial NAS and AdvRush to DO-framework, and introduce sequential finetuning: Harmonic Mean and Nash; to allow the finetuning of multi-models. Extensive experiments with real-world datasets such as MNIST, CIFAR-10, SVHN and TinyImageNet show that our variants of DONAS-GAN and DONAS-AT have significant improvements in comparison to their respective base architectures in terms of both subjective image quality and quantitative metrics.

\bibliographystyle{ieee_fullname}
\bibliography{IEEE_TIP_manuscript}

\begin{thebibliography}{10}\itemsep=-1pt

\bibitem{arora2017generalization}
Sanjeev Arora, Rong Ge, Yingyu Liang, Tengyu Ma, and Yi Zhang.
\newblock Generalization and equilibrium in generative adversarial nets ({GANs}).
\newblock In {\em ICML}, pages 224--232, 2017.

\bibitem{aung2021gan}
Aye Phyu~Phyu Aung, Xinrun Wang, Runsheng Yu, Bo An, Senthilnath Jayavelu, and Xiaoli Li.
\newblock {DO-GAN}: A double oracle framework for generative adversarial networks.
\newblock {\em CVPR}, 2022.

\bibitem{bosansky2013double}
Branislav Bo{\v{s}}ansk{\'y}, Christopher Kiekintveld, Viliam Lisy, Jiri Cermak, and Michal Pechoucek.
\newblock Double-oracle algorithm for computing an exact {Nash} equilibrium in zero-sum extensive-form games.
\newblock In {\em AAMAS}, pages 335--342, 2013.

\bibitem{bosansky2012iterative}
B Bosansky, Christopher Kiekintveld, Viliam Lisy, and Michal Pechoucek.
\newblock Iterative algorithm for solving two-player zero-sum extensive-form games with imperfect information.
\newblock In {\em ECAI}, pages 193--198, 2012.

\bibitem{brock2018large}
Andrew Brock, Jeff Donahue, and Karen Simonyan.
\newblock Large scale {GAN} training for high fidelity natural image synthesis.
\newblock In {\em ICLR}, 2018.

\bibitem{coates2011analysis}
Adam Coates, Andrew Ng, and Honglak Lee.
\newblock An analysis of single-layer networks in unsupervised feature learning.
\newblock In {\em Proceedings of the fourteenth international conference on artificial intelligence and statistics}, pages 215--223. JMLR Workshop and Conference Proceedings, 2011.

\bibitem{farnia2020gans}
Farzan Farnia and Asuman Ozdaglar.
\newblock {GANs} may have no {Nash} equilibria.
\newblock {\em arXiv preprint arXiv:2002.09124}, 2020.

\bibitem{gao2020adversarialnas}
Chen Gao, Yunpeng Chen, Si Liu, Zhenxiong Tan, and Shuicheng Yan.
\newblock {AdversarialNAS}: Adversarial neural architecture search for gans.
\newblock In {\em CVPR}, pages 5680--5689, 2020.

\bibitem{geraeinejad2021roco}
Vahid Geraeinejad, Sima Sinaei, Mehdi Modarressi, and Masoud Daneshtalab.
\newblock {RoCo-NAS}: Robust and compact neural architecture search.
\newblock In {\em IJCNN}, 2021.

\bibitem{gong2019autogan}
Xinyu Gong, Shiyu Chang, Yifan Jiang, and Zhangyang Wang.
\newblock {AutoGAN}: Neural architecture search for generative adversarial networks.
\newblock In {\em ICCV}, pages 3224--3234, 2019.

\bibitem{goodfellow2014generative}
Ian Goodfellow, Jean Pouget-Abadie, Mehdi Mirza, Bing Xu, David Warde-Farley, Sherjil Ozair, Aaron Courville, and Yoshua Bengio.
\newblock Generative adversarial nets.
\newblock In {\em NeurIPS}, pages 2672--2680, 2014.

\bibitem{goodfellow2014explaining}
Ian~J Goodfellow, Jonathon Shlens, and Christian Szegedy.
\newblock Explaining and harnessing adversarial examples.
\newblock {\em arXiv preprint arXiv:1412.6572}, 2014.

\bibitem{guo2020meets}
Minghao Guo, Yuzhe Yang, Rui Xu, Ziwei Liu, and Dahua Lin.
\newblock When {NAS} meets robustness: In search of robust architectures against adversarial attacks.
\newblock In {\em CVPR}, pages 631--640, 2020.

\bibitem{he2016deep}
Kaiming He, Xiangyu Zhang, Shaoqing Ren, and Jian Sun.
\newblock Deep residual learning for image recognition.
\newblock In {\em CVPR}, pages 770--778, 2016.

\bibitem{hoang2018mgan}
Quan Hoang, Tu~Dinh Nguyen, Trung Le, and Dinh Phung.
\newblock {MGAN}: Training generative adversarial nets with multiple generators.
\newblock In {\em ICLR}, 2018.

\bibitem{hsieh2018finding}
Ya-Ping Hsieh, Chen Liu, and Volkan Cevher.
\newblock Finding mixed nash equilibria of generative adversarial networks.
\newblock In {\em ICML}, pages 2810--2819, 2019.

\bibitem{jain2011double}
Manish Jain, Dmytro Korzhyk, Ond{\v{r}}ej Van{\v{e}}k, Vincent Conitzer, Michal P{\v{e}}chou{\v{c}}ek, and Milind Tambe.
\newblock A double oracle algorithm for zero-sum security games on graphs.
\newblock In {\em AAMAS}, pages 327--334, 2011.

\bibitem{karras2019style}
Tero Karras, Samuli Laine, and Timo Aila.
\newblock A style-based generator architecture for generative adversarial networks.
\newblock In {\em Proceedings of the IEEE/CVF Conference on Computer Vision and Pattern Recognition}, pages 4401--4410, 2019.

\bibitem{kornblith2019similarity}
Simon Kornblith, Mohammad Norouzi, Honglak Lee, and Geoffrey Hinton.
\newblock Similarity of neural network representations revisited.
\newblock In {\em International Conference on Machine Learning}, pages 3519--3529. PMLR, 2019.

\bibitem{cifarten}
Alex Krizhevsky, Vinod Nair, and Geoffrey Hinton.
\newblock {CIFAR-10 (Canadian Institute for Advanced Research)}.
\newblock https://www.cs.toronto.edu/~kriz/cifar.html, 2009.

\bibitem{lanctot2017unified}
Marc Lanctot, Vinicius Zambaldi, Audrunas Gruslys, Angeliki Lazaridou, Karl Tuyls, Julien P{\'e}rolat, David Silver, and Thore Graepel.
\newblock A unified game-theoretic approach to multiagent reinforcement learning.
\newblock In {\em NeurIPS}, pages 4190--4203, 2017.

\bibitem{Le2015TinyIV}
Ya Le and Xuan Yang.
\newblock Tiny {ImageNet} visual recognition challenge.
\newblock {\em CS 231N}, 7(7):3, 2015.

\bibitem{liu2018darts}
Hanxiao Liu, Karen Simonyan, and Yiming Yang.
\newblock {DARTS}: Differentiable architecture search.
\newblock In {\em ICLR}, 2018.

\bibitem{madry2017towards}
Aleksander Madry, Aleksandar Makelov, Ludwig Schmidt, Dimitris Tsipras, and Adrian Vladu.
\newblock Towards deep learning models resistant to adversarial attacks.
\newblock {\em arXiv preprint arXiv:1706.06083}, 2017.

\bibitem{mcmahan2003planning}
H~Brendan McMahan, Geoffrey~J Gordon, and Avrim Blum.
\newblock Planning in the presence of cost functions controlled by an adversary.
\newblock In {\em ICML}, pages 536--543, 2003.

\bibitem{mescheder2017numerics}
Lars Mescheder, Sebastian Nowozin, and Andreas Geiger.
\newblock The numerics of {GANs}.
\newblock In {\em NeurIPS}, pages 1825--1835, 2017.

\bibitem{mok2021advrush}
Jisoo Mok, Byunggook Na, Hyeokjun Choe, and Sungroh Yoon.
\newblock Advrush: Searching for adversarially robust neural architectures.
\newblock In {\em ICCV}, pages 12322--12332, 2021.

\bibitem{rezaei2021generative}
Seyed Saeed~Changiz Rezaei, Fred~X Han, Di Niu, Mohammad Salameh, Keith Mills, Shuo Lian, Wei Lu, and Shangling Jui.
\newblock Generative adversarial neural architecture search.
\newblock {\em arXiv preprint arXiv:2105.09356}, 2021.

\bibitem{shafahi2019adversarial}
Ali Shafahi, Mahyar Najibi, Amin Ghiasi, Zheng Xu, John Dickerson, Christoph Studer, Larry~S Davis, Gavin Taylor, and Tom Goldstein.
\newblock Adversarial training for free!
\newblock {\em arXiv preprint arXiv:1904.12843}, 2019.

\bibitem{tsai2012security}
Jason Tsai, Thanh~H Nguyen, and Milind Tambe.
\newblock Security games for controlling contagion.
\newblock In {\em AAAI}, pages 1464--1470, 2012.

\bibitem{varshney2020stm}
Sakshi Varshney, PK Srijith, and Vineeth~N Balasubramanian.
\newblock Stm-gan: Sequentially trained multiple generators for mitigating mode collapse.
\newblock In {\em International Conference on Neural Information Processing}, pages 676--684. Springer, 2020.

\bibitem{waugh2009strategy}
Kevin Waugh, Nolan Bard, and Michael Bowling.
\newblock Strategy grafting in extensive games.
\newblock In {\em NeurIPS}, pages 2026--2034, 2009.

\bibitem{xie2023point}
Tao Xie, Haoming Zhang, Linqi Yang, Ke Wang, Kun Dai, Ruifeng Li, and Lijun Zhao.
\newblock Point-nas: A novel neural architecture search framework for point cloud analysis.
\newblock {\em IEEE Transactions on Image Processing}, 32:6526--6542, 2023.

\bibitem{zheng2021migo}
Xiawu Zheng, Rongrong Ji, Yuhang Chen, Qiang Wang, Baochang Zhang, Jie Chen, Qixiang Ye, Feiyue Huang, and Yonghong Tian.
\newblock Migo-nas: Towards fast and generalizable neural architecture search.
\newblock {\em IEEE Transactions on Pattern Analysis and Machine Intelligence}, 43(9):2936--2952, 2021.

\end{thebibliography}

\section*{Biography}
\vspace{-1cm}
\begin{IEEEbiography}[{\includegraphics[width=1in,height=1.25in,clip,keepaspectratio]{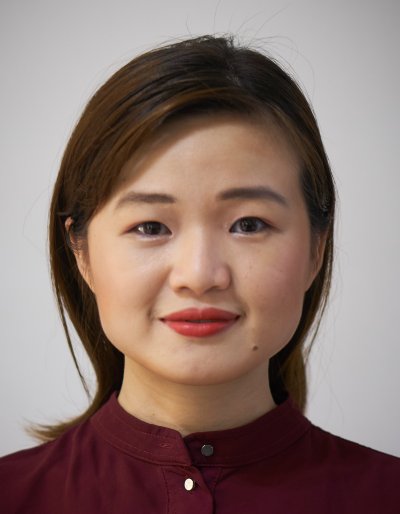}}]
{Aye Phyu Phyu Aung} earned her Ph.D. Degree in Computer Science from Nanyang Technological University, Singapore in 2023, with seminal works published in CVPR, ICAPS, and Neurocomputing. Her research interests include generative models, computational game theory, reinforcement learning and optimization. Affiliated with Institute of Infocomm Research (I2R), A*STAR, Singapore, her collaborative research works aim to bridge academia and real-world tech solutions.
\end{IEEEbiography} 

\vspace{-1cm}
\begin{IEEEbiography}
[{\includegraphics[width=1in,height=1.25in,clip,keepaspectratio]{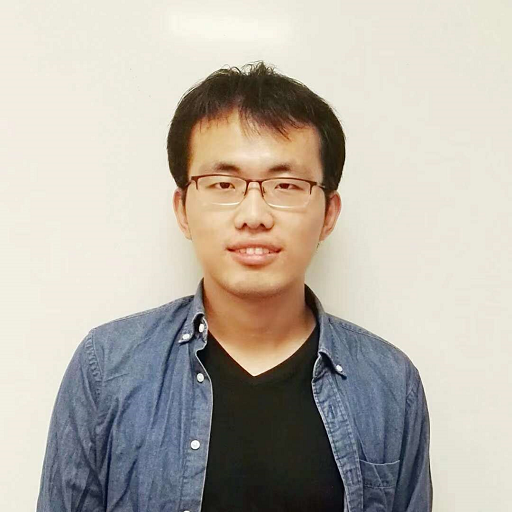}}]
    {Xinrun Wang} is a Research Fellow in Nanyang Technological University (NTU). He received the PhD degree from NTU in 2020 and the bachelor degree in Dalian University of Technology. His research interests include game theory, multi-agent reinforcement learning and distributed foundation models. 
\end{IEEEbiography}

\vspace{-1cm}
\begin{IEEEbiography}[{\includegraphics[width=1in,height=1.25in,clip,keepaspectratio]{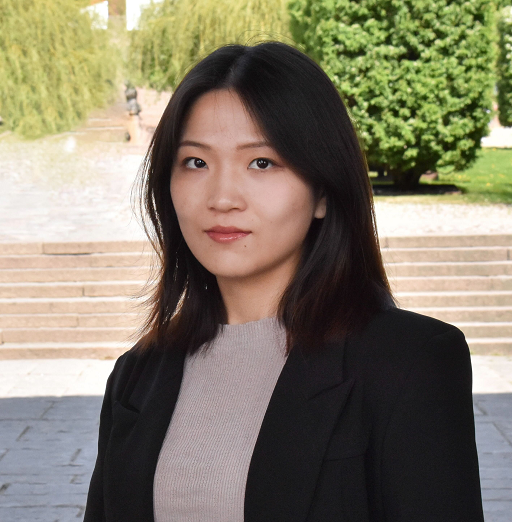}}]
{Ruiyu Wang} is currently a PhD student at Robotics, Perception, and Learning (RPL) of KTH with Prof. Florian Pokorny. My research is supported by the CloudRobotics project of WASP and my research interests fall in developing machine learning methodologies for robotic manipulation in large-scale cloud robotics system.
\end{IEEEbiography} 

\vspace{-1cm}
\begin{IEEEbiography}
[{\includegraphics[width=1in,height=1.25in,clip,keepaspectratio]{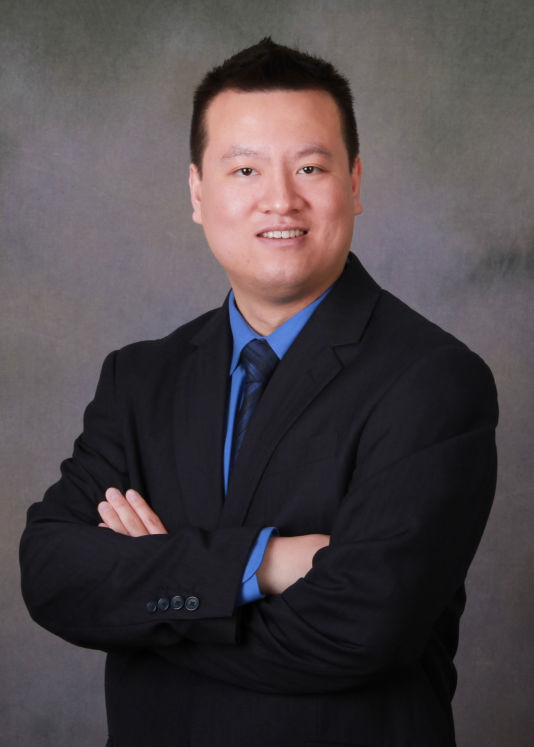}}]
    {Hau Chan} is currently an assistant professor in the School of Computing at the University of Nebraska-Lincoln. He received his Ph.D. in Computer Science from Stony Brook University in 2015 and completed three years of Postdoctoral Fellowships, including at the Laboratory for Innovation Science at Harvard University in 2018. His research aims to establish mathematical and algorithmic foundations to address societal problems from AI and optimization perspectives. He leverages computational game theory and algorithmic mechanism design when the problems involve analyzing/predicting and inducing agent strategic behavior, respectively, and ML and algorithms when the problems deal with predictions and optimizations with non-strategic agents. His research has been supported by NSF, NIH, and USCYBERCOM. He has received several Best Paper Awards at SDM and AAMAS and distinguished/outstanding SPC/PC member recognitions at IJCAI and WSDM. He has given tutorials and talks on computational game theory and mechanism design at venues such as AAMAS and IJCAI since 2018, including an Early Career Spotlight at IJCAI 2022. He has served as a co-chair for Doctoral Consortium, Scholarships, and Diversity and Inclusion Activities at AAMAS and IJCAI.
\end{IEEEbiography}

\vspace{-1cm}
\begin{IEEEbiography}
[{\includegraphics[width=1in,height=1.25in,clip,keepaspectratio]{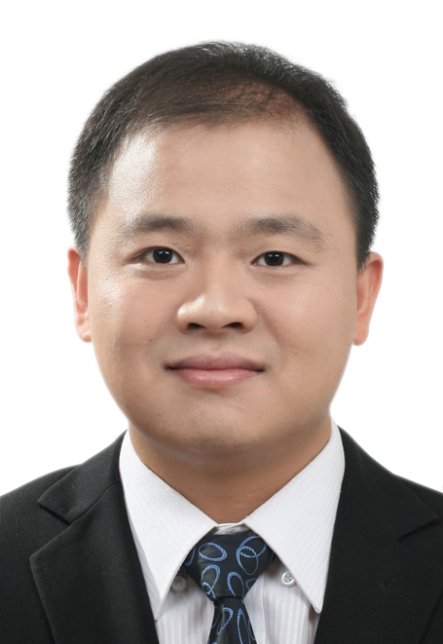}}]
{Bo An} is a President’s Council Chair Professor and Co-Director of Artificial Intelligence Research Institute (AI.R) at Nanyang Technological University, Singapore. He received the Ph.D degree in Computer Science from the University of Massachusetts, Amherst. His current research interests include artificial intelligence, multiagent systems, computational game theory, reinforcement learning, and optimization. He has published over 150 referred papers at AAMAS, IJCAI, AAAI, ICLR, NeurIPS, ICML, AISTATS, ICAPS, KDD, UAI, EC, WWW, JAAMAS, and AIJ. Dr. An was the recipient of the 2010 IFAAMAS Victor Lesser Distinguished Dissertation Award, an Operational Excellence Award from the Commander, First Coast Guard District of the United States, the 2012 INFORMS Daniel H. Wagner Prize for Excellence in Operations Research Practice, 2018 Nanyang Research Award (Young Investigator), and 2022 Nanyang Research Award. His publications won the Best Innovative Application Paper Award at AAMAS’12, the Innovative Application Award at IAAI’16, and the best paper award at DAI’20. He was invited to give Early Career Spotlight talk at IJCAI’17. He led the team HogRider which won the 2017 Microsoft Collaborative AI Challenge. He was named to IEEE Intelligent Systems' "AI's 10 to Watch" list for 2018. He was PC Co-Chair of AAMAS’20 and General Co-Chair of AAMAS’23. He will be PC Chair of IJCAI. He is a member of the editorial board of JAIR and is the Associate Editor of AIJ, JAAMAS, IEEE Intelligent Systems, ACM TAAS, and ACM TIST. He was elected to the board of directors of IFAAMAS, senior member of AAAI, and Distinguished member of ACM. 
\end{IEEEbiography}

\vspace{-1cm}
\begin{IEEEbiography}
[{\includegraphics[width=1in,height=1.25in,clip,keepaspectratio]{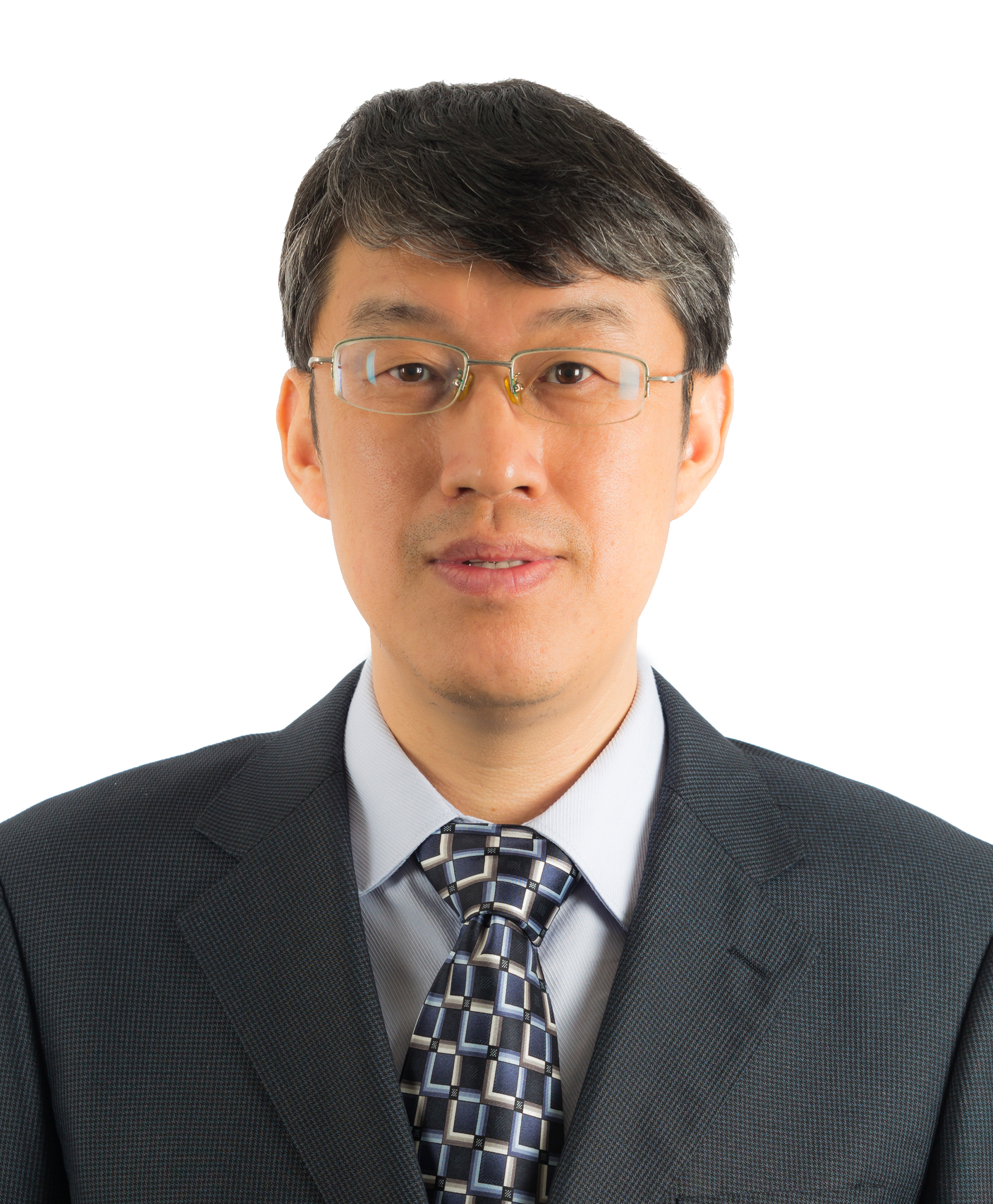}}]
    Xiaoli Li is currently a senior principal scientist and dept head at the Institute for Infocomm Research, A*STAR, Singapore. He also holds adjunct full professor position at Nanyang Technological University. His research interests include AI, machine learning, data mining, and bioinformatics. He has been serving as general chairs/area chairs/senior PC members in AI and data mining related conferences. He is currently serving as editor-in-chief of World Scientific Annual Review of Artificial Intelligence, and associate editors of IEEE Transactions on Artificial Intelligence, Knowledge and Information Systems and Machine Learning with Applications. Xiaoli has published more than 300 high quality papers and won 9 best paper/benchmark competition awards.
\end{IEEEbiography}

\vspace{-1cm}
\begin{IEEEbiography}[{\includegraphics[width=1in,height=1.25in,clip,keepaspectratio]{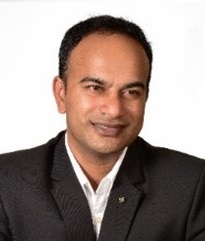}}]{J. Senthilnath} (Senior Member, IEEE) is a Senior Scientist in the Institute for Infocomm Research (I$^{2}$R) at the Agency for Science, Technology and Research (A*STAR), Singapore. He received the PhD degree in Aerospace Engineering from the Indian Institute of Science (IISc), India. His current research interests include artificial intelligence, multi-agent systems, generative models, online learning, reinforcement learning and optimization. He has published over 100 high-quality papers and won five best paper awards. He has been serving as Guest Editor/organizing chair/co-chair/PC members in leading AI and data analytics journals and conferences.
\end{IEEEbiography}

\onecolumn
\setcounter{section}{0}
\setcounter{figure}{0}
\setcounter{table}{0}
\setcounter{algocf}{0}
\setcounter{page}{1}
%\section{Comparison of Terminologies between Game Theory and GAN}
%\label{appendixTerminolgies}

% \section{Comparison of Terminologies between Game Theory and GAN}
% \label{appendixTerminolgies}
% \begin{table*}[ht]
% \centering
% \caption{Comparison of Terminologies between Game Theory and GAN}
% \begin{tabular}{ll}
% \toprule\toprule
% Game Theory terminology & GAN terminology \\ \midrule\midrule
% Player                  & Generator/ discriminator  \\ \midrule
% Strategy                & The parameter setting of generator/ discriminator, e.g., $\pi_g$ and $\pi_d$ \\\midrule
% \multirow{2}{*}{Policy}               & The sequence of parameters (strategies) till epoch $t$, e.g., ($\pi^1_g, \pi^2_g, ..., \pi^t_g$) \\
% & Note: Not used in DO-GAN.\\\midrule
% Game                    & The minmax game between generator and discriminator\\\midrule
% \multirow{2}{*}{Meta-game/ meta-matrix}  & The minmax game between generator \& discriminator with \\
% & their respective set of strategies at epoch $t$ of DO framework                             \\\midrule
% Meta-strategy           & The mixed NE strategy of generator/discriminator at epoch $t$  \\ \bottomrule\bottomrule                                    
% \end{tabular}
% \end{table*}

\section{Generated images from DONAS-GAN}
\label{appendixCDONAS}

\begin{figure}[H]
    \centering    
    \begin{subfigure}[b]{0.48\linewidth}
\includegraphics[width=\linewidth]{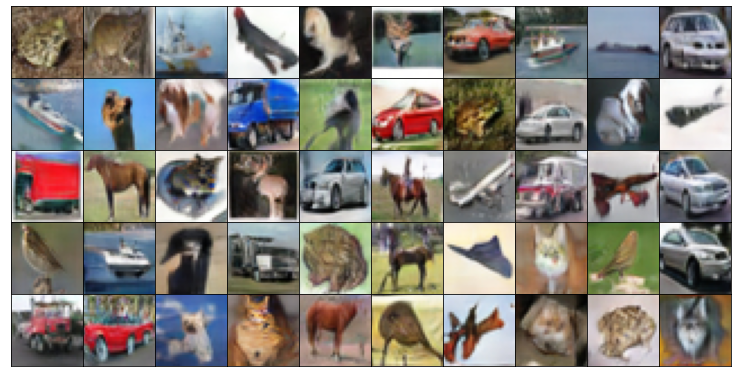}
    \caption{The CIFAR-10 images randomly generated from DONAS for GAN using $K=5$}
    \label{fig:my_sublabel}
    \end{subfigure}
    \begin{subfigure}[b]{0.48\linewidth}
    \includegraphics[width=\linewidth]{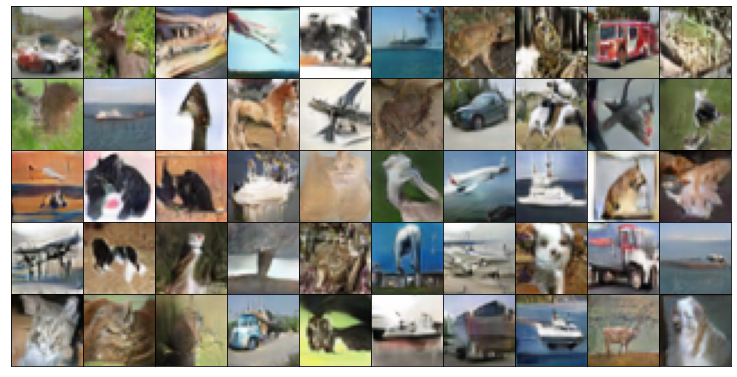}
    \caption{The CIFAR-10 images randomly generated from DONAS for GAN using $K=10$}
    \label{fig:my_label_10}
    \end{subfigure}
    \label{fig:my_label}
\end{figure}

% \begin{figure}[H]
%     \centering
    
% \end{figure}

Figure~\ref{fig:my_sublabel} and Figure~\ref{fig:my_label_10} show the realistic image generation with any mode-collapse through the generated images of DONAS-GAN $(K=5, 10)$ for CIFAR-10 dataset.

\section{Full internal representation of DONAS-GAN networks by CKA analysis}
\label{appendixEDONAS}

\noindent We use the centered kernel alignment (CKA) proposed by Kornblith et al to measure the similarity of the representation between layers and networks of DONAS-GAN trained on TinyImageNet. CKA is a commonly used metric for representation similarity, which features in higher accuracy comparing with other similarity indexes. We observe that the search process can obtain diverse architectures (e.g., Generator 0, 1, and 3) as the mixed architecture to generate images with different distribution to mitigate the mode-collapse problem.

\begin{figure}[H]
    \centering
    \includegraphics[width=0.8\linewidth]{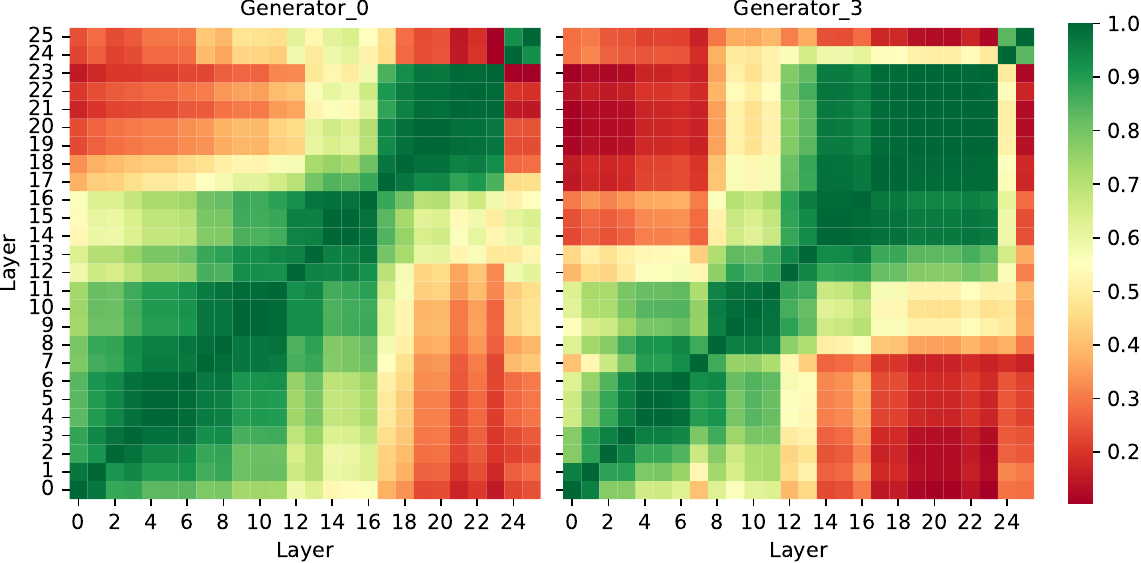}
    \caption{The search process of DONAS-GAN obtaining diverse architectures.}
    \label{fig:cka_generator_self_2}
\end{figure}

\begin{figure}[H]
    \centering
    \includegraphics[width=0.7\linewidth]{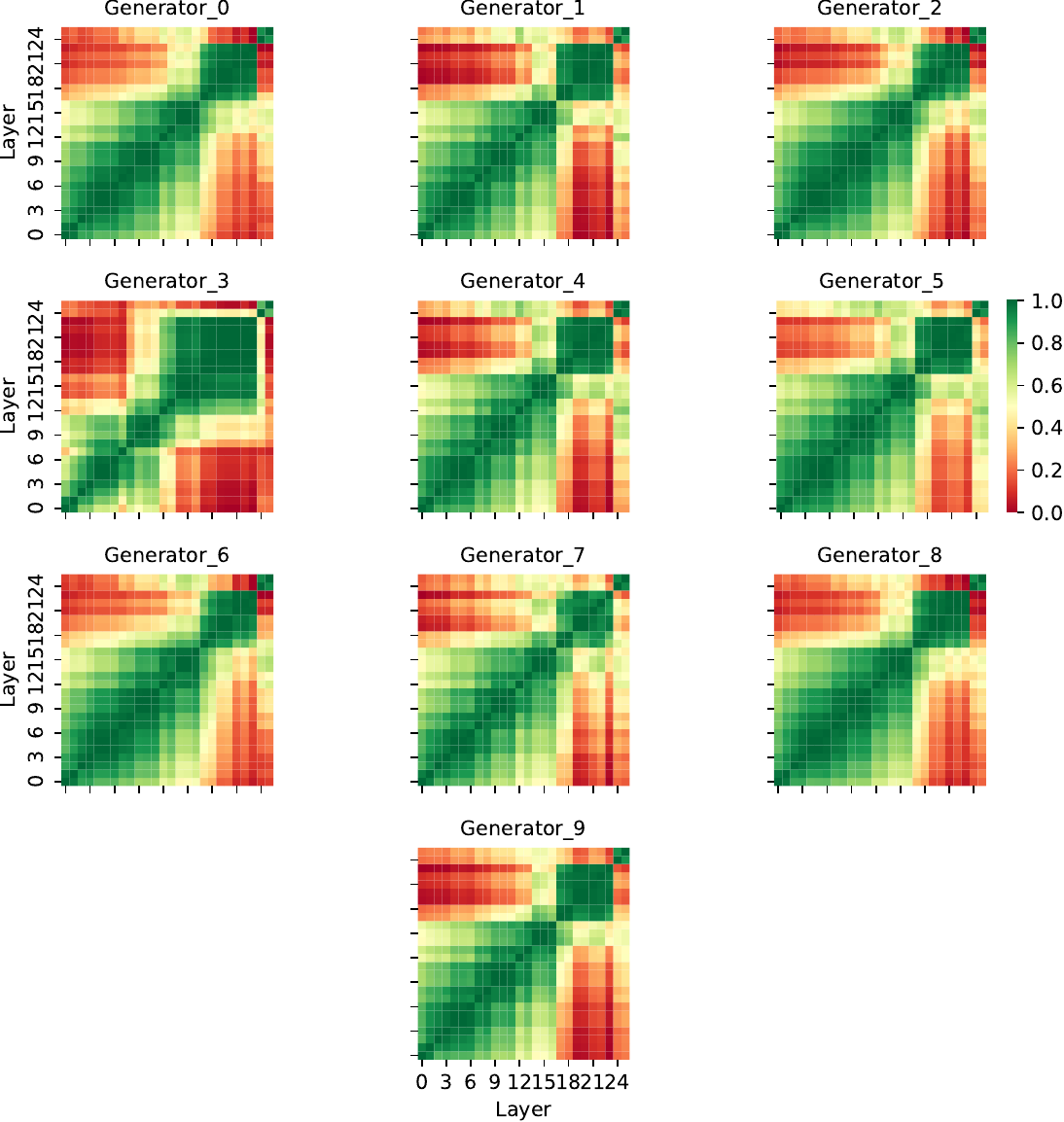}
    \caption{Linear CKA between layers of individual searched generator networks by DONAS-GAN for TinyImageNet dataset.}
    \label{fig:self_appendix}
\end{figure}

\begin{figure}[H]
    \centering
\includegraphics[width=0.7\linewidth]{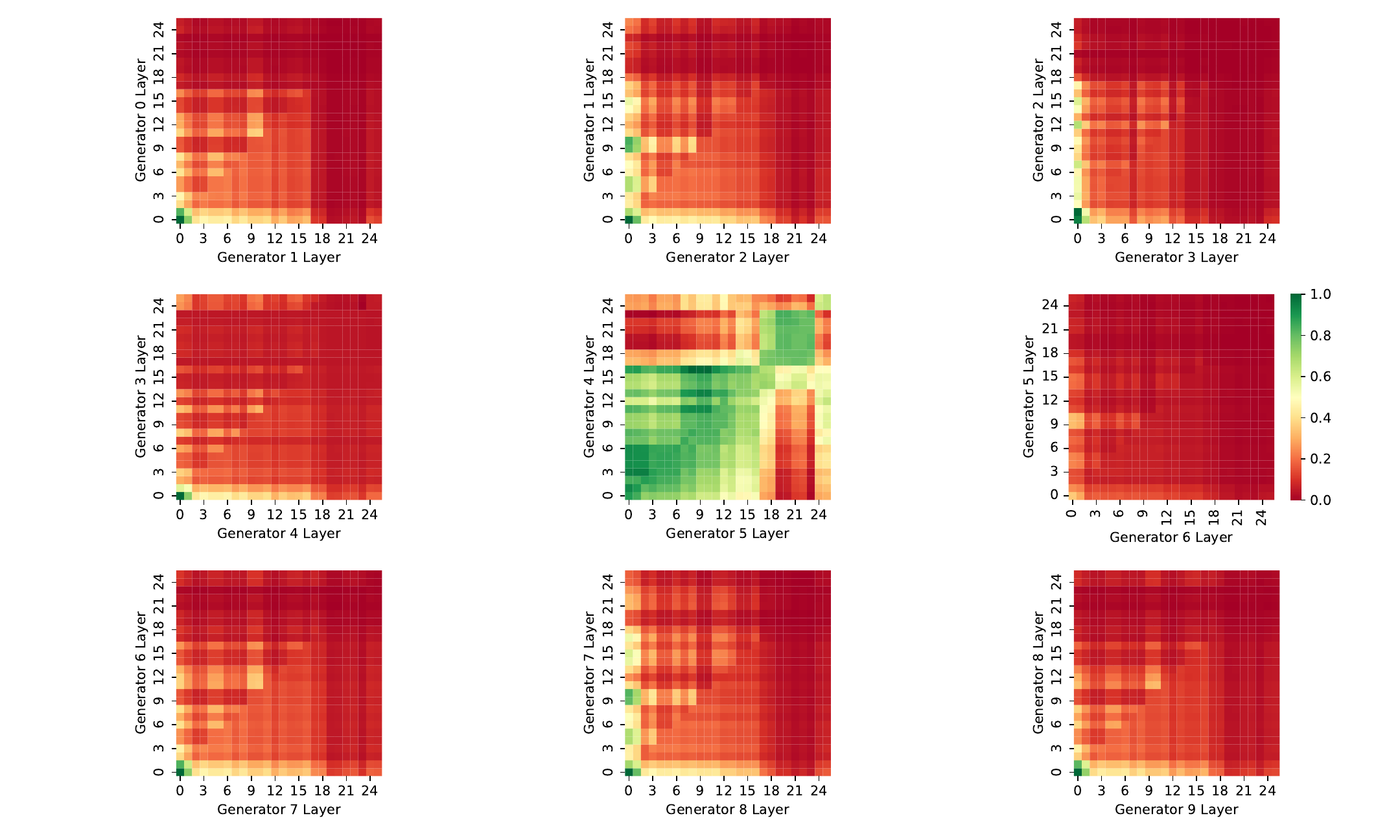}
\caption{Linear CKA between layers of searched generator networks by DONAS-GAN for TinyImageNet dataset.}
\label{fig:mutal_appendix}
\end{figure}

\end{document}